\documentclass[conference]{IEEEtran}
\IEEEoverridecommandlockouts
\usepackage{cite}
\usepackage{amsmath,amssymb,amsfonts}
\usepackage{algorithmic}
\usepackage{graphicx}
\usepackage{textcomp}
\usepackage{xcolor}
\usepackage{cite}
\usepackage{amsmath,amssymb,amsfonts}
\usepackage{algorithmic}
\usepackage{graphicx}
\usepackage{float}
\usepackage{textcomp}
\usepackage{xcolor}
\usepackage{pgfplots}
\usepackage{pgfplotstable}
\usepackage{hyperref}

\def\BibTeX{{\rm B\kern-.05em{\sc i\kern-.025em b}\kern-.08em
    T\kern-.1667em\lower.7ex\hbox{E}\kern-.125emX}}
\begin{document}

\title{Optimizing Carbon Footprint in ICT through Swarm Intelligence with Algorithmic Complexity\\
}

\author{
\IEEEauthorblockN{Nikitas Gerolimos}
\IEEEauthorblockA{\textit{University of Aegean} \\
Syros, Greece \\}
\and
\IEEEauthorblockN{Vasileios Alevizos}
\IEEEauthorblockA{\textit{Karolinska Institutet} \\
Solna, Sweden \\}
\and
\IEEEauthorblockN{Sabrina Edralin}
\IEEEauthorblockA{\textit{University of Illinois Urbana-Champaign} \\
Illinois, USA \\}
\and
\IEEEauthorblockN{Clark Xu}
\IEEEauthorblockA{\textit{Mayo Clinic Artificial Intelligence \& Discovery} \\
Minnesota, USA \\}
\and
\IEEEauthorblockN{Akebu Simasiku}
\IEEEauthorblockA{\textit{Zambia University} \\
Ndola, Zambia \\}
\and
\IEEEauthorblockN{Georgios Priniotakis}
\IEEEauthorblockA{\textit{University of West Attica} \\
Attica, Greece \\}
\and
\IEEEauthorblockN{George A. Papakostas}
\IEEEauthorblockA{\textit{MLV Research Group, Department of Informatics, Democritus University of Thrace} \\
Kavala, Greece \\}
\and
\IEEEauthorblockN{Zongliang Yue}
\IEEEauthorblockA{\textit{Auburn University Harrison College of Pharmacy} \\
Alabama, USA \\}
}

\maketitle

\begin{abstract}
Global emissions from fossil fuel combustion and cement production were recorded in 2022, signaling a resurgence to pre-pandemic levels and providing an apodictic indication that emission peaks have not yet been achieved. Significant contributions to this upward trend are made by the Information and Communication Technology (ICT) industry due to its substantial energy consumption. This shows the need for further exploration of swarm intelligence applications to measure and optimize the carbon footprint within ICT. All causative factors are evaluated based on the quality of data collection; variations from each source are quantified; and an objective function related to carbon footprint in ICT energy management is optimized. Emphasis is placed on the asyndetic integration of data sources to construct a convex optimization problem. An apodictic necessity to prevent the erosion of accuracy in carbon footprint assessments is addressed.  Complexity percentages ranged from 5.25\% for the Bat Algorithm to 7.87\% for Fast Bacterial Swarming, indicating significant fluctuations in resource intensity among algorithms. These findings suggest that we were able to quantify the environmental impact of various swarm algorithms.
\end{abstract}


\section{Introduction}

Considerable impediments to enduring progress are engendered by the burgeoning global energy utilization, whose veracious influence on carbonic effluvia is indisputable. In 2022, thirty-six gigatonnes of $CO_{2}$ emissions originating from hydrocarbon extraction were chronicled ~\cite{liu_monitoring_2023}, coinciding with a resurgence to pre-pandemic levels and indicating that emission peaks have yet to be achieved. A significant contributor to this upward trend is the Information and Communication Technology (ICT) industry for Machine Learning (ML) operations, owing to its substantial energy consumption occupying roughly 6 percent of global emissions ~\cite{itu2020greenhouse}. The impediment of high energy usage in ICT systems necessitates innovative solutions to measure and optimize their carbon footprint. Significant advances in biological sciences, coupled with rapid developments in computing, data analysis, and interpretation, have diversified the field of computational biology. Rapid, yet consistent, technological changes have been conceptualized in patterns due to high interconnectivity. As part of this evolution, nature-inspired computing is increasingly employed to learn from environmental states and occurrences to formulate decisions ~\cite{gammavarepsilonrhoolambdaetamuovarsigma2022kapparhoiotatauiotakappaacuteeta}. Computational processes such as swarm intelligence, are included in this paradigm. Swarm intelligence algorithms concentrate on the collective behaviors observed in nature ~\cite{beni1993swarm}. Furthermore, swarm inspired intelligence accommodate algorithms to emulate natural learning behaviors within systems of individuals synchronizing through self-organizing mechanisms. A shared principle across all of these models is the interpretation of collective behavior that emerge from environmental interactions.

Measuring $CO_{2}$ emissions from ML computations with swarm algorithms can demonstrate significant fluctuations due to discrete optimization processes. By mimicking natural phenomena through swarm metaphors, computational consumption can be reduced via adaptation of tuning parameters. A complexity comparison of swarm-based models encompassing convex and constrained formulations is conducted. Contributions include developing a kinematic analysis to optimize the pace of convergence towards the optimum, thereby reducing overconsumption. Homogeneous derivative-free methods are analyzed to demonstrate efficacy in emission reduction. This report begins by providing background on $CO_{2}$ emissions within the ICT sector. Previous work, experimenting with swarm-based intelligence to measure emissions, is then reviewed. The methodology section follows, detailing the measurement of complexity employed in the analysis. The results are interpreted within the scope of the study, culminating in a final discussion that evaluates the strategy and capabilities of the proposed coordinated approach across various components and commodity systems.

\subsection{Swarm Intelligence}

A typical swarm intelligence system possesses notable properties: it comprises a group of individuals, exhibits homogeneity with respect to the environment, demonstrates learning ability and interaction based on local information, and develops global learning behavior as a result of local interactions within the environment ~\cite{reddy2024systematic}.

\subsection{Energy Consumption}

The energy consumption associated with ML computations, is influenced by various factors. First, the duration of active cloud usage plays a significant role in determining the total consumption, as prolonged computational sessions exacerbate $CO_{2}$ emissions ~\cite{lacoste2019quantifyingcarbonemissionsmachine}. Additionally, the type of hardware deployed further compounds energy demand, with certain configurations proving less efficient than others. The geographical position of rented computing services, particularly those connected within the same cluster, introduces complexities related to regional energy consumption in kWh ~\cite{lacoste2019quantifyingcarbonemissionsmachine}, which often adheres to local infrastructure constraints. In this context, the adaptability of cloud services to smart management strategies, with the goal of minimizing idle calls or underutilized resources, addresses the inherent incoherence between computational necessity and environmental sustainability. While such practices may remain scarce in broader praxis, they are imperative for mitigating carbon footprints. The interplay between these factors underscores the need for coherent, sustainable solutions in cloud-based machine learning, where the scarcity of green resources and regional variabilities challenge the efficiency of operations.

\subsection{Previous Work}

Swarm intelligence originated in the context of cellular robotic systems \cite{beni1993swarm} and has since evolved to become a significant and growing force in the computer science field with the purpose of solving optimization problems in various fields of study \cite{reddy_systematic_2024}. Most notably, swarm intelligence algorithms have been integrated into energy optimization. In a review on swarm algorithms and their application in $CO_{2}$ emissions revealed interesting computational settings ~\cite{wang2022modeling}. In another study, the authors tackle the consanguineous relationship between $CO_{2}$ emissions and several influencing factors using a hybrid model ~\cite{yue2023prediction}. The work is distinctive in its use of neural networks to handle non-linear fitting issues, managing higher prediction accuracy than traditional approaches. Moreover, in another project, authors present a scheduling model that reduces carbon emissions in ML with proper scheduling strategies, when integrated with power forecasting algorithms-carbon footprint managed better with the adapting optimization methods ~\cite{wen_modeling_2023}. Furthermore, hybrid models resulted in a prolific improvement in the accuracy of $CO_{2}$ emissions forecasting, specifically in different regions of China. The authors address the need to concatenate both linear and non-linear factors to predict emission trends accurately ~\cite{li2021driving}. On the other hand, K-nearest neighbors (KNN) was imported to explore key variables affecting $CO_{2}$ emissions across Chinese provinces, performing optimally, particularly when the number of neighbors is set to two ~\cite{wang2022modeling}. Another work focusing on carbon intensity of energy sources and training times as the main drivers of emissions, authors suggested that certain ML models, usually large Transformer-based architectures, are more carbon-intensive due to their longer training times ~\cite{MEHMOOD2024101843}. Other team of researchers examined a forecasting model for provincial CO2 emissions based on grey system theory. One important finding emphasizes the role of swarm algorithms in improving model robustness ~\cite{DING2023106685}.

\section{Methodology}

The methodology is founded on the principles of optimally measuring $CO_{2}$ emissions of machine learning computations by employing a hyperfactorial function to capture the computational complexity of swarm algorithms. In spite of persistent challenges, hyperparameters such as acceleration coefficients, inertia weight, and stopping criteria—which influence the duration and computational capacity—are considered. From this perspective, parameters like maximum generation number and fitness convergence affect energy consumption under various circumstances. Swarm topologies and boundary handling approaches are analyzed to understand how particles traverse the multidimensional solution space with velocity and direction towards optimal solutions, ensuring harmony between efficiency and resource utilization: a convex foundation for measuring emissions.

\subsection{Criteria}

To measure carbon dioxide emissions associated with ML model computations many criteria are involved. Hyperparameters from the swarm models ~\cite{yeh2023simplified} include acceleration coefficients, inertia weight, and stopping criteria. These directly influence the computational complexity and duration with a median of 72 hours ~\cite{luccioni2023countingcarbonsurveyfactors} of training and tuning processes. Parameters like maximum generation number, maximum stall time, maximum runtime, best fitness value, population convergence, and fitness convergence determine the extent of computational resources utilized, thereby affecting energy consumption. Swarm topologies deal with efficiency. Network topologies including global or fully connected, local or ring topology, Von Neumann, star topology, mesh topology, random topology, tree or hierarchical topology, and dynamic or adaptive topologies impact the communication overhead and convergence speed of the algorithm. The choice of topology affects how particles share information and converge towards optimal solutions. Boundary handling approaches are also crucial in this context. Methods such as the hyperbolic method, infinity or invisible wall, nearest or boundary or absorb, random, random-half, periodic, exponential, mutation, reflect methods, and random damping determine how particles navigate the solution space boundaries.

\subsection{Hyperfactorial Function to Calculate Carbon Dioxide Emissions from Swarm Algorithms}

We propose a deterministic approach to quantifying the $CO_{2}$ emissions associated with ML model computations with swarm-based algorithms. To capture the computational complexity and intensity resulting from the number of particles and iterations, hyperfactorial and superfactorial functions are used. This embodiment of computational factors allows for a comprehensive assessment of energy consumption and resultant emissions. Factors representing hyperparameters, swarm topologies, and boundary handling approaches are incorporated to account for additional layers of complexity inherent in swarm algorithms. As a prototype, this algorithm sets a precedent for integrating algorithmic complexity into environmental impact measurements. Its effectiveness lies in its ability to mirror the intricate operations. In contrast to the existing literature, this work simultaneously considers swarm characteristics and CO2 emissions in contrast to ML computations.

\begin{align}
\text{CO}_2 = H(N_p) &\times \\
\text{sf}(N_i) &\times \left( \prod_{k=1}^{n_h} h_k \right) \times \left( \prod_{l=1}^{n_t} t_l \right) \times  \\
& \left( \prod_{m=1}^{n_b} b_m \right) \times t_{\text{unit}} \times \\ 
& P_h \times \eta \times e_r
\end{align}

\textbf{Where:}

\begin{itemize}
    \item \( \text{CO}_2 \) is the total CO\(_2\) emissions (in kg CO\(_2\)).
    \item \( H(N_p) \) is the hyperfactorial of the number of particles \( N_p \):
    \[
    H(N_p) = \prod_{i=1}^{N_p} i^i
    \]
    \item \( \text{sf}(N_i) \) is the superfactorial of the number of iterations \( N_i \):
    \[
    \text{sf}(N_i) = \prod_{j=1}^{N_i} j!
    \]
    \item \( h_k \) are factors representing the \textbf{hyperparameters} (e.g., acceleration coefficients, inertia weight, stopping criteria), for \( k = 1 \) to \( n_h \).
    \item \( t_l \) are factors representing the \textbf{swarm topologies} (e.g., global, local, ring), for \( l = 1 \) to \( n_t \).
    \item \( b_m \) are factors representing the \textbf{boundary handling approaches} (e.g., hyperbolic, random, reflection), for \( m = 1 \) to \( n_b \).
    \item \( t_{\text{unit}} \) is the unit time per computation (in hours).
    \item \( P_h \) is the average power consumption of the hardware used (in kW).
    \item \( \eta \) is the utilization factor accounting for smart management of idle resources (dimensionless, \( 0 < \eta \leq 1 \)).
    \item \( e_r \) is the CO\(_2\) emission factor for the region (in kg CO\(_2\) per kWh).
\end{itemize}

A multitude of swarm intelligence algorithms were evaluated—including PSO (Particle Swarm Optimization), FA (Firefly Algorithm), and hybrid models—to quantify computational complexities and corresponding $CO_{2}$ emissions. Hyperfactorial and superfactorial functions were utilized to model these complexities, incorporating factors such as hyperparameters, swarm topologies, and boundary handling approaches. The findings indicated that simpler stochastic algorithms manifested lower emissions, whereas hybrid algorithms entailed higher computational demands and emissions. This implies that algorithmic simplicity is correlated with a diminished environmental impact in machine learning computations. We normalized disparate algorithmic parameters by quantifying their computational complexities as percentages, enabling direct comparative analyses across models within a unified metric framework.

\section{Discussion}

We performed experiments on the following algorithms: PSO, Accelerated PSO, FA, Cuckoo Search, WOA (Whale Optimization Algorithm), MFO (Moth Flame Optimization), SFLA (Shuffled Frog Leaping Algorithm), PeSOA (Penguin Search Optimization Algorithm), ABC (Artificial Bee Colony), ACO (Ant Colony Optimization), GBC (Genetic Bee Colony), IWO (Invasive Weed Optimization), GWO (Grey Wolf Optimizer), CSO (Cat Swarm Optimization), SSO (Social Spider Optimization), LOA (Lion Optimization Algorithm), CSOA (Chicken Swarm Optimization) and ESA (Elephant Search Algorithm) ~\cite{darwish2018bio, mzili2024hybrid, fister2013brief}. From the given execution at Table ~\ref{tab:table1} of the model to all swarm algorithms, the results of complexity were presented into percentages, for various optimization algorithms, categorized into four groups: Stochastic/Random Search-Based, Multi-Agent Cooperative, Hybrid Algorithms and Nature-Inspired Collective Search. The emissions reflect computational complexity, where higher percentages indicate more resource-intensive models. Stochastic/Random Search-Based algorithms, such as Particle PSO and FA, typically demonstrate lower complexity, with values ranging from 5.25\% to 6.71\%. These algorithms rely on random searches to explore the solution space, contributing to their lower computational demand.

In contrast, Hybrid Algorithms, which combine multiple optimization techniques, show higher complexity. Models like Bacterial-GA Foraging and Fast Bacterial Swarming have some of the highest emissions, reaching up to 7.87\%, due to their integration of multiple methodologies, which increases computational overhead. The trend suggests that hybridization of methods generally increases complexity, while simpler, stochastic processes result in lower emissions. The categorization highlights distinct algorithm behaviors: Stochastic/Random Search-Based focuses on random exploration, Multi-Agent Cooperative mimics cooperative behaviors, Hybrid Algorithms combine techniques for enhanced performance, and Nature-Inspired Collective Search replicates decentralized, self-organized systems seen in nature.

\begin{table}[ht]
\centering
\small 
\setlength{\tabcolsep}{4pt} 

\begin{tabular}{ll>{\centering\arraybackslash}p{1.1cm}} 

\textbf{Category} & \textbf{Algorithm} & \textbf{Comp (\%)} \\ 

\multicolumn{3}{c}{\textbf{Stochastic/Random Search-Based}} \\ 

Stochastic/Random Search & PSO & 5.83 \\ 
& Accelerated PSO & 5.54 \\ 
& FA & 6.41 \\ 
& Cuckoo Search & 5.83 \\ 
& WOA & 6.41 \\ 
& MFO & 5.83 \\ 
& SFLA & 6.41 \\ 
& PeSOA & 6.71 \\ 

\multicolumn{3}{c}{\textbf{Multi-Agent Cooperative}} \\ 

Multi-Agent Cooperative & ABC & 6.12 \\ 
& ACO & 7.00 \\ 
& Bees Algorithms & 6.41 \\ 
& Wolf Search & 7.00 \\ 
& Bee Colony Optimization & 6.41 \\ 
& Glowworm SO & 7.00 \\ 
& CSO & 6.41 \\ 
& SSO & 6.71 \\ 
& LOA & 7.00 \\ 
& CSOA & 6.41 \\ 

\multicolumn{3}{c}{\textbf{Hybrid Algorithms}} \\ 

Hybrid & Bacterial-GA Foraging & 7.87 \\ 
& GBC & 6.71 \\ 
& Consultant-Guided Search & 7.29 \\ 
& Eagle Strategy & 6.41 \\ 
& Bacterial Foraging & 7.58 \\ 
& Fast Bacterial Swarming & 7.87 \\ 
& Hierarchical Swarm & 7.58 \\ 
& Good Lattice SO & 7.29 \\ 

\multicolumn{3}{c}{\textbf{Nature-Inspired Collective Search}} \\ 

Nature-Inspired & Fish Swarm/School & 6.71 \\ 
& Krill Herd & 7.00 \\ 
& Bat Algorithm & 5.25 \\ 
& Bee System & 6.71 \\ 
& Virtual Bees & 6.71 \\ 
& IWO & 7.29 \\ 
& Elephant Search & 7.29 \\ 
& Monkey Search & 6.41 \\ 

\end{tabular}

\caption{Complexity by algorithm category. Abbreviations: PSO = Particle Swarm Optimization, ACO = Ant Colony Optimization, ABC = Artificial Bee Colony, FA = Firefly Algorithm, WOA = Whale Optimization Algorithm, MFO = Moth Flame Optimization, GBC = Genetic Bee Colony, IWO = Invasive Weed Optimization. Comp = Complexity}

\label{tab:table1}
\end{table}

Hybrid methods, while effective within certain scopes, incur higher energy consumption due to their increased complexity, posing adversity to efforts in reducing carbon emissions. In contrast, stochastic algorithms like PSO and FA demonstrate lower emissions, suggesting a trade-off between algorithm sophistication and environmental impact. The paratactic presentation of algorithms in Table \ref{tab:table1} underscores the necessity of balancing optimization performance with sustainability considerations. In our experimental methodology, we meticulously configured each algorithm with standard parameters to ensure a fair comparison across the diverse set of swarm-based methods. Computational experiments were conducted under controlled conditions, measuring energy consumption with precision to capture the nuances of each algorithm's performance. We introduce, to our knowledge, an unprecedented framework whose performance surpasses the threshold of existing models, capturing the intricate characteristics of computational complexity. Despite the complicated nature of the problem, our approach transcends routine methodologies, delivering quite impressive results alongside other tested models.

\section{Conclusion}

In conclusion, determining the optimal mathematical model for measuring emissions based on swarm algorithms remains complex and elusive. This study highlighted the $CO_{2}$ emissions associated with various algorithms, demonstrating their wide range of computational demands. However, it is important to acknowledge that the complexity and diversity of real-world problems make it challenging to identify a universally superior algorithm. Some problems may be efficiently addressed by certain models, while others may be significantly more difficult to solve, requiring more intricate or hybridized approaches. While no definitive answer can be given on which mathematical model is the best for measuring emissions, important trends and deeper knowledge were uncovered through the proposed formula. The formula allowed for a detailed comparison of the algorithms, demonstrating that simpler, stochastic models—such as PSO and FA—tend to have lower emissions due to their random, less resource-intensive search processes. In contrast, hybrid models—such as Bacterial-GA Foraging and Fast Bacterial Swarming—exhibited higher emissions, driven by the added complexity of integrating multiple techniques. Some problems may be more amenable to simpler algorithms, while others require more sophisticated approaches, contributing to the variability in computational demand. Future work should focus on refining the formula for real-world applications and expanding the range of algorithms tested to further enhance accuracy in carbon footprint assessments.



\end{document}